\documentclass[letterpaper]{article} 
\usepackage{aaai21}  
\usepackage{times}  
\usepackage{helvet} 
\usepackage{courier}  
\usepackage[hyphens]{url}  
\usepackage{graphicx} 
\usepackage{natbib}  
\usepackage{caption} 
\title{SKATE: A Natural Language Interface for Encoding Structured Knowledge }
\author{Clifton McFate,
Aditya Kalyanpur,
Dave Ferrucci, \\
Andrea Bradshaw,
Ariel Diertani, 
David Melville, 
Lori Moon \\}
\affiliations{
   Elemental Cognition \\
   New York, NY, USA \\
   \{cjm, adityak, davef, andreab, arield, davidm, lorim\}@elementalcognition.com \\ 

}
\date{August 2020}

\begin{document}

\maketitle

\begin{abstract}
\noindent In Natural Language (NL) applications, there is often a mismatch between what the NL interface is capable of interpreting and what a lay user knows how to express. This work describes a novel natural language interface that reduces this mismatch by refining natural language input through successive, automatically generated semi-structured templates. In this paper we describe how our approach, called SKATE, uses a neural semantic parser to parse NL input and suggest semi-structured templates, which are recursively filled to produce fully structured interpretations. We also show how SKATE integrates with a neural rule-generation model to interactively suggest and acquire commonsense knowledge. We provide a preliminary coverage analysis of SKATE for the task of story understanding, and then describe a current business use-case of the tool in a specific domain: COVID-19 policy design.
\end{abstract}

\section{Introduction}
Interactive natural language applications typically require mapping spoken or written language to a semi-formal structure, often represented using semantic frames with fillable slots. This approach has been used in popular commercial spoken dialogue systems (e.g., Google's Dialogflow and Amazon's Alexa skills) through the developer-defined ``intents.'' Frame semantic parsing more broadly \citep[e.g.,][]{gildea2002automatic}  has demonstrated benefit in a number of downstream applications including dialogue systems \citep{chen2013unsupervised} and question answering \citep{shen2007using}.

Despite advances in frame semantic parsing  \citep[e.g.,][]{swayamdipta2017frame}, no semantic parser is perfect. Accordingly, developers of natural language interfaces must carefully curate correction dialogues to avoid frustrating interactions. This sort of mismatch between system and user expectations is what we aim to resolve with SKATE (Structured Knowledge AcquisiTion and Extraction).


In SKATE, a user's text is parsed in real time as they type. The resulting partial semantic structures can be completed with additional required slots and fillers, and are then recursively refined by the user through micro-dialogues. At any point, the user can continue to give structured interpretations for a slot filler (e.g., a complex noun phrase), or they can leave it in unstructured form for the system to interpret later.

In the following sections, we first walk through the SKATE architecture using an exemplar interaction from an open-domain rule learning task. We then summarize our implementation of the core SKATE components. We demonstrate how SKATE has been integrated with a natural language rule generation model to interactively acquire structured rules for story understanding, and conclude with a current application that uses SKATE to build COVID-19 policy diagrams.

\begin{figure*}[t]
\centering
\includegraphics[width=0.8\textwidth]{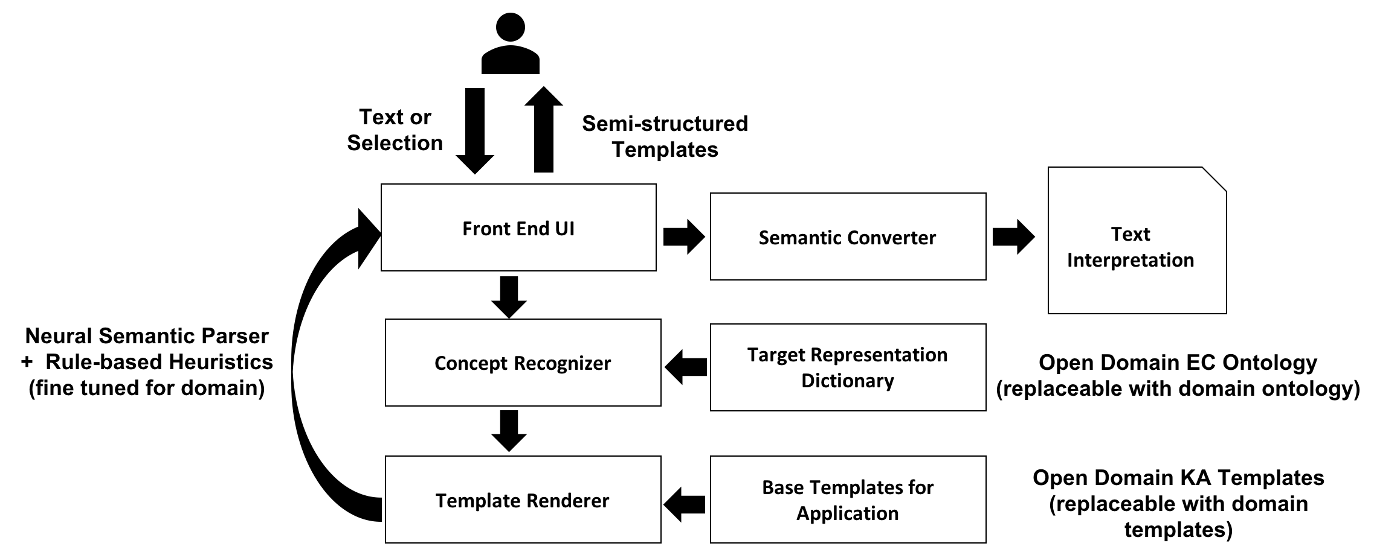}
\caption{SKATE Architecture Diagram}.
\label{fig:arch}
\end{figure*}

\section{SKATE Overview}
The SKATE architecture (Figure \ref{fig:arch}) is built around an interaction model of recursively: recognizing a concept, producing a partially interpreted template to instantiate the concept, and allowing a user to refine the template. The result is text annotated with semantic frames. These frames are processed by the downstream application.

\begin{figure}
\centering
\includegraphics[width=.9\columnwidth]{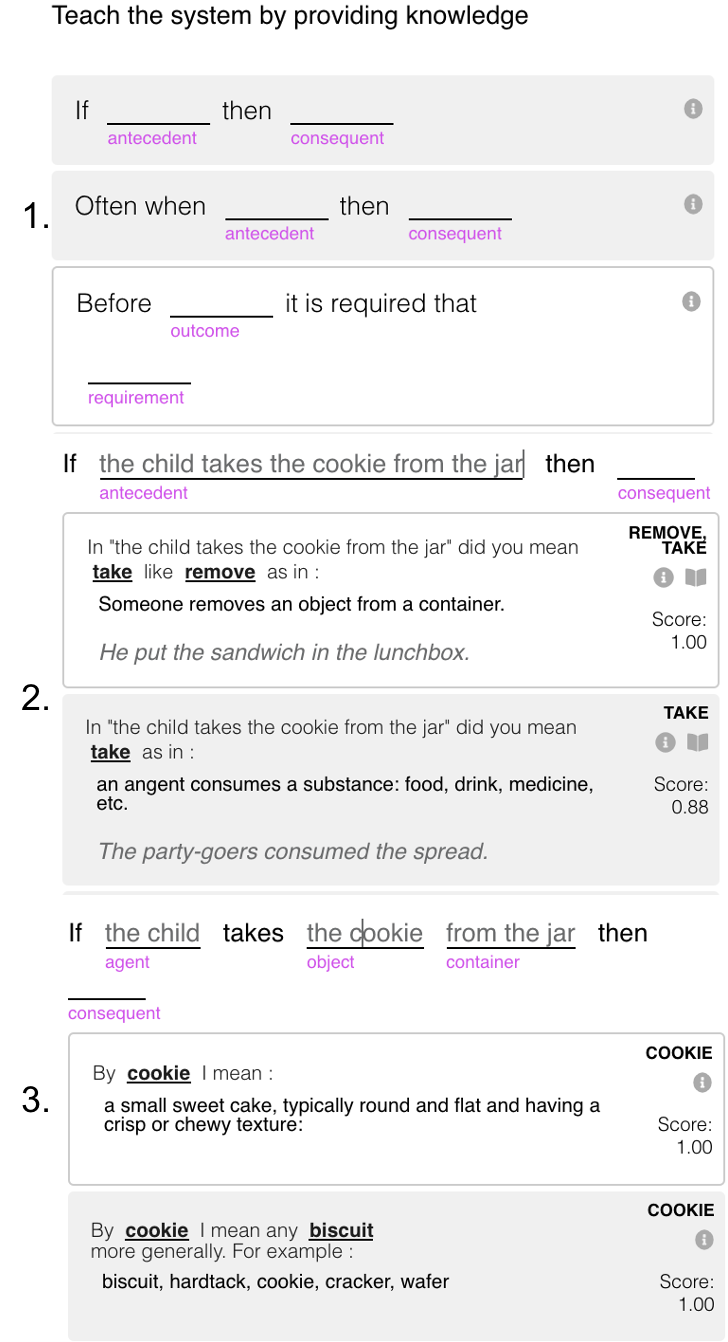}
\caption{Example SKATE flow for entering ``The child takes the cookie from the jar.''}
\label{fig:cookie-flow}
\end{figure}

Each interaction begins by selecting a top-level, application-specific semantic template. As an example, a top-level frame for rule acquisition may be an ``If/Then'' construction as in the first pane of Figure \ref{fig:cookie-flow}. These top-level templates provide the initial scope of interaction, and can be used to apply additional application-specific semantics as needed (e.g. If/Then could produce a causal rule while After/Then might only imply temporal sequence).

As a user fills a slot, the concept recognizer processes their text, selects a lexical trigger, and instantiates possible semantic frames for that trigger. The frames are in the style of FrameNet \citep{ruppenhofer2006framenet}: each defines a concept, a set of possible trigger phrases, and semantic arguments that may be instantiated as text spans. At each interaction, we use syntactic heuristics to select the trigger with the widest syntactic scope.

For each instantiated frame, the template renderer receives the interpretations (frame predicates and optionally argument labels/spans) from the concept recognizer and decorates that information (e.g., by adding display texts, examples etc) to send to the front-end UI. It also presents the user with options for what frame to assign as the word sense of the trigger. For example, in the second pane of Figure \ref{fig:cookie-flow}, the template generator has built frame assignment options for the word ``take.'' The resulting micro-dialogue is presented to the user. Once an option is selected, the corresponding template is displayed, and the user can recursively refine unstructured slot fillers as in the third pane of Figure \ref{fig:cookie-flow}. The user can also choose to leave slots as unstructured text. For instance, in Figure \ref{fig:cookie-flow}, the user may not need to specify the desired sense of ``cookie,'', and the entry can be submitted without full specification (in which case, uninterpreted tokens become placeholders/variables in the underlying semantic representation, and can be refined at a later point).

Note that in many domains, it is necessary to solicit extra information from a user given an evoked frame. When instantiating templates, required roles that remain unfilled can be added to the template to appear as blank slots for the user to specify (and must be filled in before the user submits). Additionally, likely roles suggested by context can be added and optionally deleted. 

Once a user is satisfied with what they have typed, they can submit the entry. The set of composed frames can be further processed by the application-specific semantic converter if necessary. For example, in the rule application and COVID-19 policy builder, the resulting frames are turned into a set of Horn clause-like statements. 

\section{SKATE Components}
In this section, we briefly describe the core SKATE components\footnote{Due to lack of space, we omit details for the Template Renderer which primarily sets up the UI look \& feel.}.

\subsection{Target Representation Vocabulary}
Our concept vocabulary is organized such that each predicate corresponds to a frame. All frames minimally possess the ``focal'' role which corresponds to the lexical trigger for the frame, though they may have additional optional and required roles. Frames are stored in an inheritance hierarchy, allowing multiple inheritance.

As a domain general starting point, we have created a frame ontology called Hector, derived from FrameNet \citep{ruppenhofer2006framenet} and the New  Oxford American Dictionary \citep[NOAD]{noad}. These two resources are complementary: FrameNet has broad coverage for multi-arity relations, while NOAD has a large library of lexical concepts (entities, attributes, etc.). The Hector ontology can easily be pruned into subsets for specific domains and/or expanded with novel concepts. Defining a new frame requires, minimally, defining its roles, writing a short definition or example, and optionally positioning it in the existing frame hierarchy. SKATE's performance improves with annotated examples, but they are not required, and as discussed in the next subsection, SKATE can generate its own training data as a new frame is selected by the user and elaborated upon in SKATE interactions.


\subsection{Concept Recognizer}
The concept recognizer component consists of two semantic parsers. The first, SPINDLE  \citep{Spindle}, is a transformer-based neural semantic parser. This model can be fine-tuned using a corpus, but requires annotated data. The second parser acts as a fallback and is used when Spindle returns no results or low confidence frame interpretations. It is based on an unsupervised approach that retrieves $k$ nearest frames based on an embedding match between the sentence typed so far and potential frame embeddings, the latter being generated from minimal frame annotated examples pre-specified by a domain author\footnote{This feature supports quick domain adaptation when there isn't sufficient training data for a supervised semantic parsing model.}. As SKATE is used in an application, the corrected output of the second parser becomes training data to improve the first. Thus, SKATE is able to improve with use.

\subsubsection{Supervised Frame Semantic Parsing (SPINDLE)}

\begin{figure}[h]
\centering
\includegraphics[width=1\columnwidth]{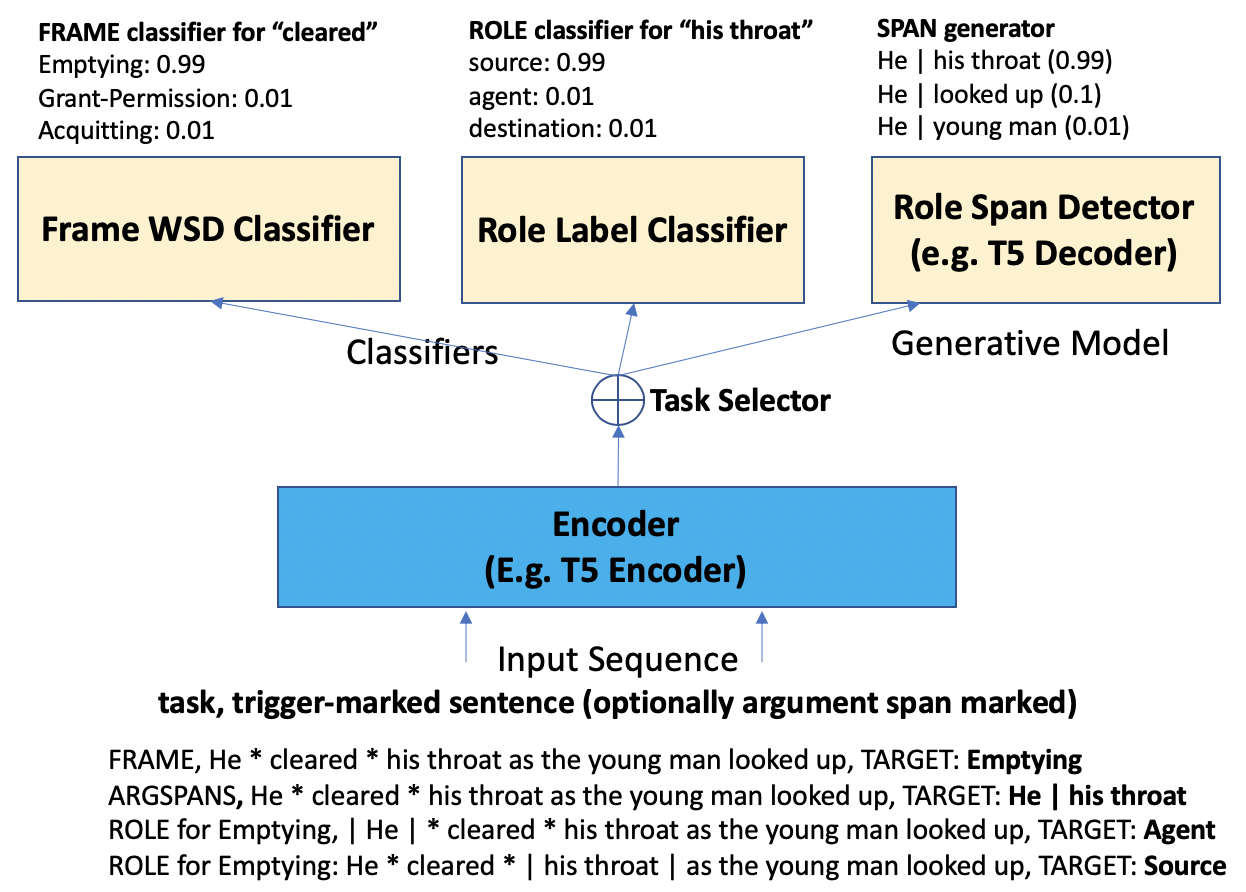}
\caption{SPINDLE: Joint multi-task encoder-decoder model for semantic parsing}
\label{fig:spindle-arch}
\end{figure}

We have developed a neural semantic parser called SPINDLE that treats frame parsing as a multi-task problem involving related classification and generation tasks. Given a sentence and a frame-triggering span, the model decomposes parsing into frame-sense disambiguation (multi-label classification), argument span detection (generation), and role-labeling (classification). Since these tasks are related, SPINDLE uses a joint multi-task encoder-decoder architecture (see Figure \ref{fig:spindle-arch}), where the encoder layer is shared among the various tasks, with different decoders used depending on the task type. 

The model is trained on ~500K annotated frame sentences (available in FrameNet and NOAD) by fine-tuning a pre-trained, transformer-based language model such as GPT2 \citep{gpt2} or T5 \citep{2019t5}. The SPINDLE model achieved the best results using T5 as the base encoder/decoder, with a frame sense disambiguation accuracy of 91\% and a span detection/role labeling F1 of 84\%. Even though the parser was trained on full sentences, we have found that it returns results with high accuracy when run on partial sentences like those typed in SKATE. Moreover, as the user continues to type text, the parsing results change to consider the additional context, which helps to disambiguate the correct frame sense. 

\begin{figure*}[t]
\centering
\includegraphics[width=0.8\textwidth]{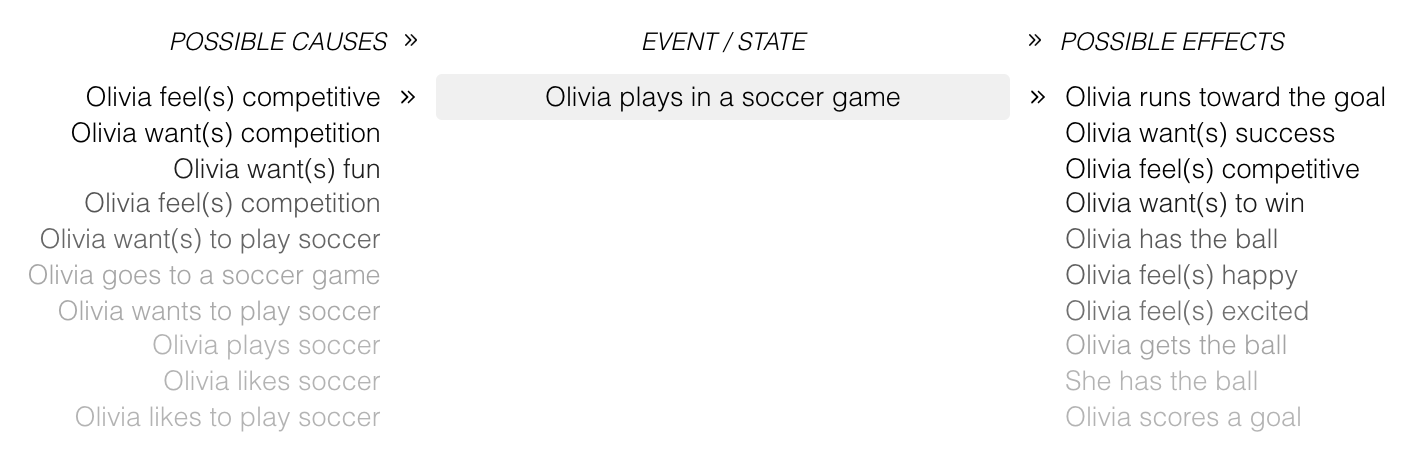}
\caption{GLUCOSE suggestions generated from a children's story about soccer}
\label{fig-glucose1}
\end{figure*}

\subsubsection{Embedding-Based Heuristic Parsing}
To complement the neural semantic parser, which needs many annotated examples for training, we have developed an unsupervised, $k$-NN-based approach for frame parsing that can work with a handful of examples per frame. The approach first computes a frame embedding by aggregating GLoVe \citep{pennington2014glove} embeddings for trigger lemmas (which are specified in the frame definition) and content words in frame examples. Our tool then sums the GLoVe embeddings for all words in the sentence to produce a sentence embedding, and computes the similarity between the frame and sentence embeddings. The algorithm also detects argument spans using syntactic heuristics based on a dependency parse of the sentence. Finally, it assigns a role for each span by considering how well the type of the span phrase matches the expected role type as inferred from frame examples (type similarity checking is also done using embeddings).  

\subsection{Semantic Converter}
The result of a submitted entry is a possibly incomplete frame-semantic parse of the input text. The semantic converter can also contain domain-specific logic to further convert the frame semantic interpretation into usable data for a downstream application (as described in the Domain Adaptation section).

\section{Commonsense Knowledge Acquisition for Story Understanding}

SKATE has been applied for open-domain structured rule acquisition. The task is: given a short story and a question, provide a rule or set of rules with which the answer can be derived from the story. Using SKATE, we can collect structured formal rules usable by a downstream reasoning engine.

As described above, SKATE templates are meant to guide the user both with explicit structure (e.g., slots) and, optionally, with unstructured slot-fillers. These unstructured fillers can be used to guide the user to submissions with high-confidence semantic parses or towards prototypical examples. For this task we integrate SKATE with a neural unstructured rule prediction system to guide the user towards general, syntactically simple, rules.

\subsection{GLUCOSE: Contextual Rule Prediction}
GLUCOSE \citep[GeneraLized and COntextualized Story Explanations;][]{glucose} is a crowd-sourced dataset of common-sense explanatory knowledge. GLUCOSE defines ten dimensions of causal explanation, focusing on events, states, motivations,  emotions,  and naive psychology. The GLUCOSE dataset consists of both general and specific semi-structured inference rules that apply to short children's stories. These rules were acquired via crowd-sourcing, and \citet{glucose} demonstrated that neural models trained on these semi-structured rules could be used to produce human-like inferences for story understanding.

\begin{figure}[h]
\includegraphics[width=0.9\columnwidth]{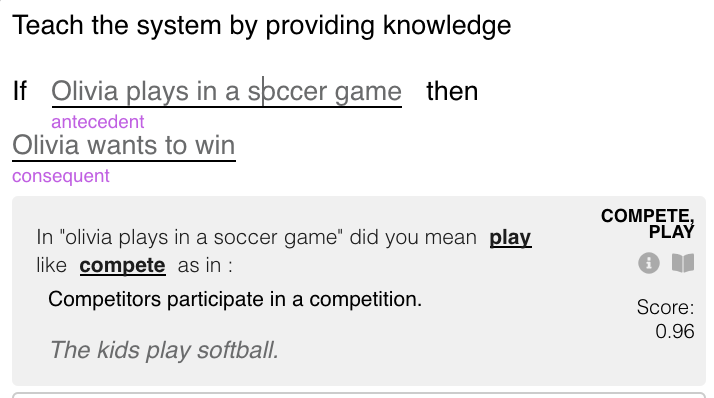}
\caption{GLUCOSE suggestion for a story about soccer.}
\label{fig:glucose2}
\end{figure}

Following \citet{glucose}, we train an encoder-decoder rule generation model. For each sentence in a story, we use the GLUCOSE trained model to predict unstructured textual causal inferences. These uninterpreted inferences are then used to seed slots in SKATE rule templates, guiding the user towards high-likelihood story-relevant rules (see Figure \ref{fig:glucose2}).

The GLUCOSE-trained model can also be used for \emph{auto-complete} suggestions in SKATE. As the user types text in one of the structured template slots, we run the model on the text typed so far (i.e., in earlier slots of the template) and generate potential completions. 
A novel feature of SKATE is that we use the already specified frame semantics to filter out incompatible language model suggestions. For example, say the user is providing knowledge about a soccer story, and starts typing: \emph{``If a player gets''} and specifies the interpretation for the verb ``get'' as the frame \emph{arriving-at-a-location}.  At this point, the frame template has an unfilled slot for ``destination''. Suppose the user continues by typing text in this slot, and we use the GLUCOSE model to generate completions, it may produce the following alternatives: \emph{``..a ball''}, \emph{``..to the goal''}, \emph{``..into trouble''} given the prior text \emph{``If a player gets''}. However, because the user has specified the frame semantics for ``get'' and the active slot is ``destination'', the only compatible suggestion is \emph{``..to the goal''}. To identify compatible suggestions, we run the SPINDLE semantic parser on the full generated completion (including the prior text) and filter out suggestions where the frame doesn't match the prior specified frame. In the above example, we would throw out \emph{``gets a ball''} (where get means \emph{acquire}), and \emph{``gets into trouble''} (where get means \emph{transition-to-state}), since it does not match the earlier specified interpretation of ``get'' (\emph{arrive-at-location}). We believe that suggesting text completions that are consistent valid semantic interpretations given prior context is unique to SKATE. 

\section{Coverage Analysis}
As a preliminary coverage evaluation, we asked domain experts to use the tool to encode knowledge needed to answer and explain commonsense questions generated from children's stories. The questions and required rules were generated in English as a part of several manually created story understanding rubrics \citep{Dunietz2020ToTM}. Our data set consisted of 340 target natural language rules from 11 children's stories. Rules could range in complexity from simple attributive statements (e.g. ``often, a house has a yard.'') to complex script-like statements (e.g. ``If a person plays soccer and the person belongs to a team and the person moves the ball to the goal then the team gets a point.'').

To test declarative statements (factoids), we additionally asked the annotators to enter, exactly or as a paraphrase, 67 sentences from an additional 4 stories.

Annotators were trained on how to use the SKATE interface and then, for each rule or statement, they rated how close in intended meaning their resulting entry was to the original NL expression on a scale from 0-3 (0 = not close; 1 = substantial deviation; 2 = minor deviation; 3 = paraphrase).

Results were promising, with 85\% of entries scoring 2 or higher, including several complex constructions involving nested clauses, conjunctions and negation. Some high scoring examples are shown in Table \ref{table:skateBKExample}. The main gaps were missing frames from the target ontology (Hector).


\begin{table}[h]
    \centering
    \small
    \begin{tabular}{|p{3.4cm}|p{4cm}|}
        \hline
        \textbf{Knowledge Target} & \textbf{SKATE Input (Score)} \\\hline
       People generally want to eat food that is tasty & Often people want to eat tasty food (3) \\\hline
       When a larger animal approaches a smaller animal, the smaller animal might get afraid & Often when animal1 approaches animal2 and size of animal1 is greater than size of animal2, animal2 feels fear (3) \\ \hline
        When one person helps another, the person being helped thanks the helper & Often when person1 helps person2, then person2 thanks person1 (3) \\ \hline
        If something is not obscured behind another object, it can be seen & If object1 does not cover object2, then someone can see object2. (2) \\ \hline
        If someone doesn't know something, and someone else tells them, then they know what it is & If person1 does not know a fact and person2 tells person1 the fact, then person1 learns the fact (3) \\ \hline
    \end{tabular}
    \caption{Commonsense rules acquired via SKATE}
    \label{table:skateBKExample}
\end{table}

\subsection{Annotation Results and Feedback}

SKATE was used by 5 domain / subject matter experts (SMEs) to specify knowledge on the story understanding task, and as mentioned above, we collected 400+ structured logical statements (rules and facts) via the tool. Note that the alternate to using SKATE, and the prior methodology used by the SMEs, was to hand-code the formal knowledge in a verbose representational syntax (similar to Prolog), which was tedious, error-prone and time-consuming. 

The SMEs impressions can be summarized as follows: 
\begin{itemize}
\item The tool is dramatically faster than writing rules in logical syntax not just because the user is writing in natural language, but also because the tool actively coaxes the desired interpretation while moving along (instead of landing up in a undesired/confused state and backtracking). 
\item The user does not need to worry about special character formatting (commas, parentheticals etc) which need to be specified correctly when encoding logical formulae.
\item The dynamic, organized nature of the information-flow to the user, such as options regarding word senses and phrase interpretations, ensures that the rules the user produces are compatible with the resources with which they are intended to interact. In traditional approaches, writing the same rules would often require a user to check the syntax against a set of rules and the word meanings against a sense ontology as separate steps. 
\end{itemize}

\subsection{Path to Deployment}

\begin{figure}[h]
\centering
\includegraphics[width=1\columnwidth]{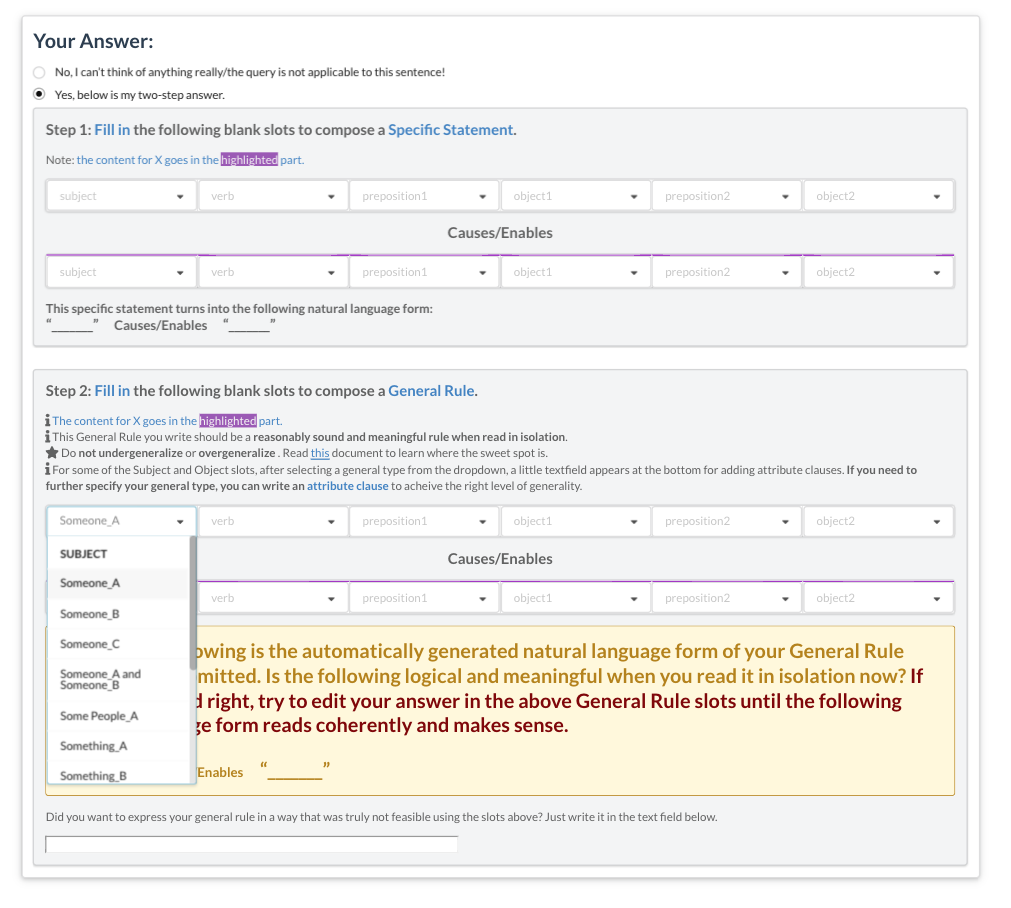}
\caption{Current GLUCOSE UI for collecting causal explanatory rules}
\label{fig:glucoseSVO}
\end{figure}

As a next step for SKATE on story understanding, we plan to use it to collect formal causal explanatory knowledge via crowdsourcing, as an extension of our previous work, GLUCOSE. The current version has a ``SKATE-lite" UI with subject-verb-object (SVO) templates (see Figure \ref{fig:glucoseSVO}). 

By using SKATE, we should be able to collect richer, more expressive knowledge, and our coverage analysis, done on the same domain and a similar/related task, has shown that the tool is more than adequate for this problem. We are keen on this deployment as it would expose SKATE to hundreds of our trained turkers.

\section{Domain Adaptation: COVID-19 Policy Diagram}
As the world recovers from COVID-19, many institutions have been required to define robust facility access policies. These policies can be complicated, often with many branching conditions (e.g. lists of symptoms) and potential actions for a user to complete (e.g. various policy compliant COVID tests). 

Automated systems can help guide users through these policies, but the policies must first be formalized. In the following section, we present an application of the SKATE NLI for building domain-specific policy diagrams around access to school facilities. 

\subsection{Task: Defining Compliance Paths}

\begin{figure*}[ht]
\centering
\includegraphics[width=0.8\textwidth]{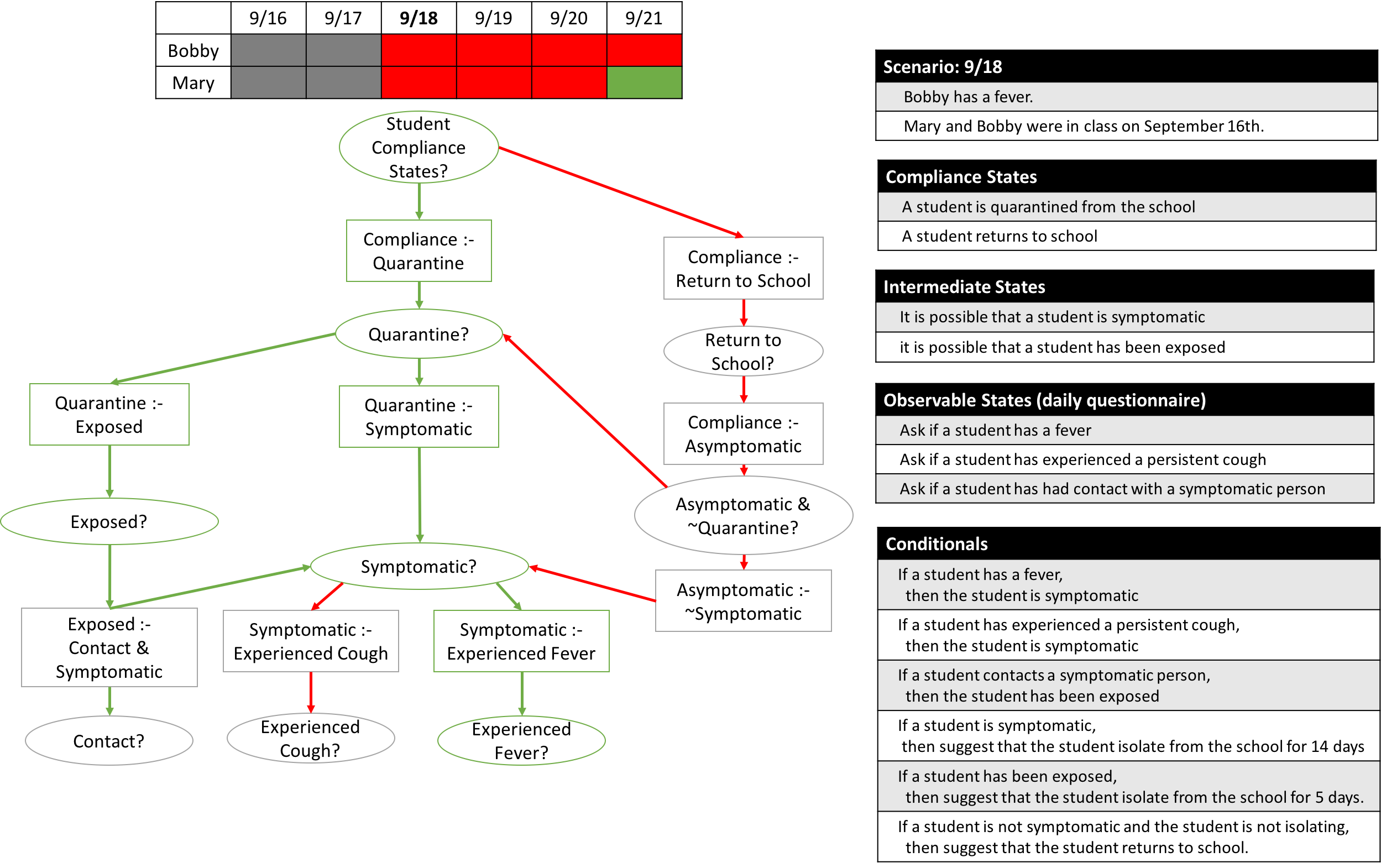}
\caption{An example policy pertaining to school access. Queries can be issued against the graph given a world state to determine compliance. Here, a query for compliance states reveals two students currently under quarantine}
\label{fig-covidEx}
\end{figure*}

A policy diagram is defined by
\begin{itemize}
\item Compliance states: terminals actions, whether a person returns or quarantines.

\item  Intermediate states: States that lead to compliance states or further modify them, e.g. quarantining because a student is symptomatic.

\item  Scenarios: observable states that lead to an intermediate state, e.g. A student experienced a cough and a fever.

\item  Variables: observable from the world, e.g. a person marked on a questionnaire that they experienced a cough.
\end{itemize}

\noindent Together these form a flow chart (policy diagram). Nodes in the diagram are states, and each type of state (above) is assigned a top-level template to allow a user to define them.

Compliance and intermediate states can be mapped to a unique frame instance or combination of frames which allows for compositionality (e.g. quarantine for 14 days / quarantine at home).  Variables are also compositional (e.g. has a persistent cough) and can be inferred by the system using rules or observed directly through an end-user questionnaire. An example of acquiring and applying a policy diagram is shown below.

\subsection{Example}

Figure \ref{fig-covidEx} shows a simplified COVID policy for returning to school along with the SKATE statements used to construct the policy.  
In this example, \emph{quarantining (from school)} and \emph{returning (to school)} have been defined as compliance states. 
Other rules append \emph{adjuncts} (optional roles) to a state (e.g. duration) when it is evoked to further specify it. Thus, we can define conditions that lead to 5 or 14 day quarantines based on whether the student was exposed or symptomatic respectively.

\begin{figure}[h]
\includegraphics[width=0.75\columnwidth]{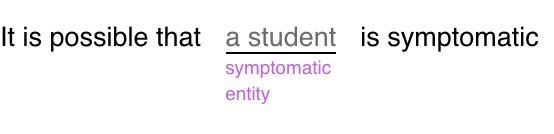}
\caption{Defining an intermediate state}
\label{stateDef.png}
\end{figure}

Figure \ref{stateDef.png} is an example entry defining an intermediate state. Later conditionals can elaborate these states (e.g. symptomatic if feverish) and target them.

\begin{figure}[h]
\includegraphics[width=0.8\columnwidth]{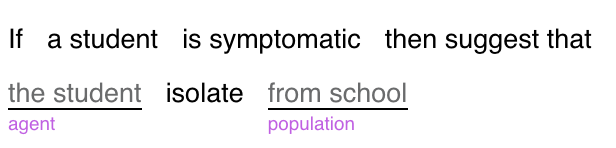}
\caption{Defining a branch of a policy diagram}
\label{complianceDef.png}
\end{figure}

Figure \ref{complianceDef.png} is an example entry defining a suggested compliance state given an intermediate state. The state is compositional, specifying a population to quarantine from. These conditionals form rules usable by a reasoning system.

We also define intermediate states (\emph{exposed} and \emph{symptomatic}) to intuitively provide reasons for suggesting a quarantine. These hold given combinations of observable facts, which can be set through a daily questionnaire.

\begin{figure*}[h]
\centering
\includegraphics[width=0.7\textwidth]{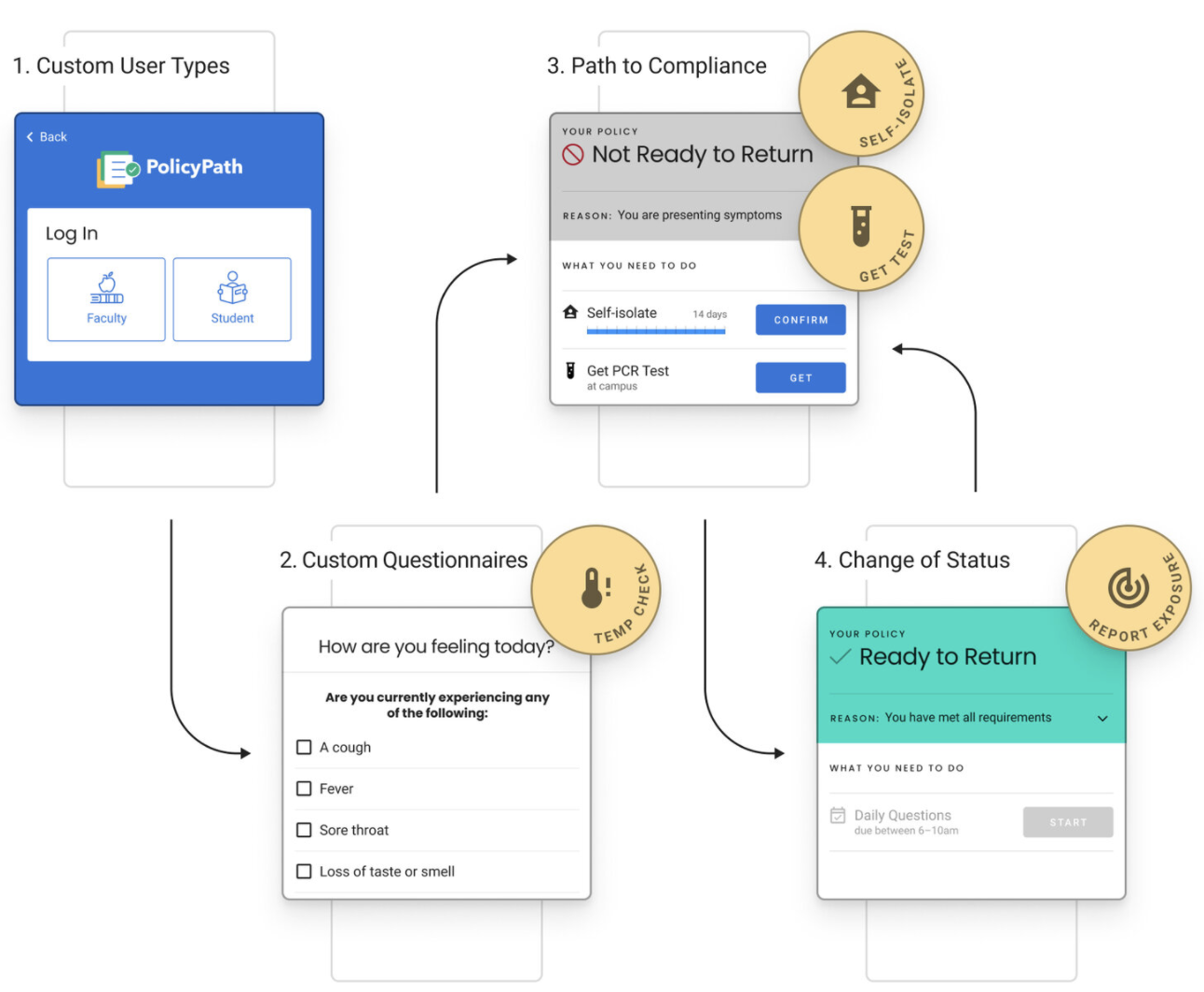}
\caption{Policy-Path Mobile App for COVID-19}
\label{fig:PolicyPaths}
\end{figure*}

In our representation, \emph{duration} adjuncts on states map to counters which can align to a calendar. Thus, we can chart when a student will be able to return to school.

The world state (i.e. a specific scenario) is also defined in SKATE (as shown in the Figure). Interestingly, in this example, "Mary and Bobby were in class.." is interpreted as \emph{co-location} in SKATE, which is used to infer \emph{contact} between the two via background knowledge in the ontology.

Given a world state (defined in SKATE), an administrator can query the graph to determine which students are in which compliance states. In the example, the system correctly infers that Bobby has 14 days left to quarantine on 9/18, while Mary only has 3 days left.

\subsection{Path to Deployment}

A team at our company, Elemental Cognition, has built a mobile app called \emph{PolicyPath} for customized and interactive COVID policy guidance. Details of the app are available at https://www.policypath.app/covid-19. 

The application, currently in use by several customers, allows an organization to define and tailor their COVID policy, which specifies the conditions that employees must satisfy before they are allowed to return to the workplace. It supports the ability to specify a custom questionnaire, testing requirements, quarantine periods, temperature checks, and more, while also providing real-time monitoring and alerts (see Figure \ref{fig:PolicyPaths}). 

In the current setup, the policy definition is typically done by the organization's management (e.g. HR department), often in a Word document, and subsequently, the logic is captured in procedural code by knowledge engineers and software developers. Our goal is to replace this process by using SKATE, as described above. The policy writer can use SKATE to express the policy in natural language, which is automatically converted to a declarative logic form under the covers. We can then use our logical reasoning and constraint checking engine called Braid \cite{kalyanpur2020braid} to enforce the semantics of the policy, when applied to a specific individual (or a group of individuals). We believe it is much easier to define, alter and test the policy when done in this ``no code" declarative paradigm.

\section{Related Work}
Natural language knowledge capture has long been a goal in AI, and interest has only grown with the advent of crowd sourcing platforms like Amazon's MTurk. Our approach draws inspiration from and improves upon this research.

ConceptNet \cite{speer2016conceptnet} started as the Open Mind Common Sense crowd-sourcing effort \cite{singh2002open} which solicited natural language common sense statements. While the OMCS knowledge acquisition interface could make use of semi-structured templates, their captured knowledge remains as natural language and they do not further decompose an entry into semantic forms. Their approach additionally used generated natural language inferences for user feedback. This plays a similar role to our auto-complete feature, though their feedback is presented after the fact rather than as inline guidance.

LEARNER \cite{chklovski2003learner} uses cumulative analogies to gather new information from ConceptNet like statements (e.g. newspapers have pages) via answerable questions (e.g. do books also have pages?). LEARNER2 builds on that design by adding templates with slots for a small set of target top-level relations \cite{chklovski2005designing}. They also generate slots to enumerate an entry, however, much like OMCS, they do not further refine input text with templates.

Our approach leverages recent advances in language modeling to generate templates from user text and to provide unstructured guidance. Recently \cite{gopinath2020fast} presented a ``contextual auto complete'' approach for clinical documentation which used a completion mechanism to disambiguate clinical concepts and create annotated notes. In contrast, our completion mechanism (templates and unstructured text) is far broader in scope (interpreting the full text) and depth of representation (compositional frames).

\section{Conclusion and Future Work}
While great advances will continue to be made in the field of semantic parsing, it is highly unlikely that any parser will always perform perfectly. As such, even when a natural language application is capable of a desired behavior, lay users face uncertainty and obstruction when their requests are wrongly interpreted.

SKATE is a Natural Language Interface that reduces the mismatch between system ability and lay user expectation by interactively guiding them towards a structured representation. Our approach combines frame-based KR with a hybrid semantic parsing approach to construct interpretations with both structured (i.e. template slots) and unstructured (i.e. textual slot fillers) content. A novel aspect of the hybrid parsing approach is its potential to automatically improve with use, since the unsupervised embedding based parser acts as a vehicle to collect training data for the supervised model. Furthermore, the use of a neural rule generation model to produce semantically valid auto-completions is a novel and significant feature from a usability standpoint.

We have demonstrated the utility of the SKATE NLI in both an open domain task (story understanding) and in a highly specialized domain (building policy diagrams). We plan to host a public endpoint demonstrating SKATE shortly.

Many challenges still remain as we integrate SKATE into an end-user application, e.g., we are exploring ways to allow users to create new frames and/or slots on the fly, when the pre-defined vocabulary is insufficient. 

\bibliography{bibliography}

\begin{thebibliography}{18}
\providecommand{\natexlab}[1]{#1}
\providecommand{\url}[1]{\texttt{#1}}
\providecommand{\urlprefix}{URL }
\expandafter\ifx\csname urlstyle\endcsname\relax
  \providecommand{\doi}[1]{doi:\discretionary{}{}{}#1}\else
  \providecommand{\doi}{doi:\discretionary{}{}{}\begingroup
  \urlstyle{rm}\Url}\fi

\bibitem[{Chen, Wang, and Rudnicky(2013)}]{chen2013unsupervised}
Chen, Y.-N.; Wang, W.~Y.; and Rudnicky, A.~I. 2013.
\newblock Unsupervised induction and filling of semantic slots for spoken
  dialogue systems using frame-semantic parsing.
\newblock In \emph{2013 IEEE Workshop on Automatic Speech Recognition and
  Understanding}, 120--125. IEEE.

\bibitem[{Chklovski(2003)}]{chklovski2003learner}
Chklovski, T. 2003.
\newblock Learner: a system for acquiring commonsense knowledge by analogy.
\newblock In \emph{Proceedings of the 2nd international conference on Knowledge
  capture}, 4--12.

\bibitem[{Chklovski(2005)}]{chklovski2005designing}
Chklovski, T. 2005.
\newblock Designing interfaces for guided collection of knowledge about
  everyday objects from volunteers.
\newblock In \emph{Proceedings of the 10th international conference on
  Intelligent user interfaces}, 311--313.

\bibitem[{Dunietz et~al.(2020)Dunietz, Burnham, Bharadwaj, Chu-Carroll, Rambow,
  and Ferrucci}]{Dunietz2020ToTM}
Dunietz, J.; Burnham, G.; Bharadwaj, A.; Chu-Carroll, J.; Rambow, O.; and
  Ferrucci, D. 2020.
\newblock To Test Machine Comprehension, Start by Defining Comprehension.
\newblock \emph{ACL2020 Theme Track} abs/2005.01525.

\bibitem[{Gildea and Jurafsky(2002)}]{gildea2002automatic}
Gildea, D.; and Jurafsky, D. 2002.
\newblock Automatic labeling of semantic roles.
\newblock \emph{Computational linguistics} 28(3): 245--288.

\bibitem[{Gopinath et~al.(2020)Gopinath, Agrawal, Murray, Horng, Karger, and
  Sontag}]{gopinath2020fast}
Gopinath, D.; Agrawal, M.; Murray, L.; Horng, S.; Karger, D.; and Sontag, D.
  2020.
\newblock Fast, Structured Clinical Documentation via Contextual Autocomplete.
\newblock \emph{arXiv preprint arXiv:2007.15153} .

\bibitem[{Kalyanpur et~al.(2020{\natexlab{a}})Kalyanpur, Biran, Breloff,
  Chu-Carroll, Diertani, Rambow, and Sammons}]{Spindle}
Kalyanpur, A.; Biran, O.; Breloff, T.; Chu-Carroll, J.; Diertani, A.; Rambow,
  O.; and Sammons, M. 2020{\natexlab{a}}.
\newblock Open-Domain Frame Semantic Parsing Using Transformers .

\bibitem[{Kalyanpur et~al.(2020{\natexlab{b}})Kalyanpur, Breloff, Ferrucci,
  Lally, and Jantos}]{kalyanpur2020braid}
Kalyanpur, A.; Breloff, T.; Ferrucci, D.; Lally, A.; and Jantos, J.
  2020{\natexlab{b}}.
\newblock Braid: Weaving Symbolic and Statistical Knowledge into Coherent
  Logical Explanations.
\newblock \emph{arXiv e-prints 2011.13354} .

\bibitem[{Mostafazadeh et~al.(2020)Mostafazadeh, Kalyanpur, Moon,
  David~Buchanan, Biran, and Chu-Carroll}]{glucose}
Mostafazadeh, N.; Kalyanpur, A.; Moon, L.; David~Buchanan, L.~B.; Biran, O.;
  and Chu-Carroll, J. 2020.
\newblock GLUCOSE: GeneraLized and COntextualized Story Explanations.
\newblock \emph{EMNLP 2020} .

\bibitem[{Pennington, Socher, and Manning(2014)}]{pennington2014glove}
Pennington, J.; Socher, R.; and Manning, C.~D. 2014.
\newblock Glove: Global vectors for word representation.
\newblock In \emph{Proceedings of the 2014 conference on empirical methods in
  natural language processing (EMNLP)}, 1532--1543.

\bibitem[{Radford et~al.(2019)Radford, Wu, Child, Luan, Amodei, and
  Sutskever}]{gpt2}
Radford, A.; Wu, J.; Child, R.; Luan, D.; Amodei, D.; and Sutskever, I. 2019.
\newblock Language Models are Unsupervised Multitask Learners .

\bibitem[{Raffel et~al.(2019)Raffel, Shazeer, Roberts, Lee, Narang, Matena,
  Zhou, Li, and Liu}]{2019t5}
Raffel, C.; Shazeer, N.; Roberts, A.; Lee, K.; Narang, S.; Matena, M.; Zhou,
  Y.; Li, W.; and Liu, P.~J. 2019.
\newblock Exploring the Limits of Transfer Learning with a Unified Text-to-Text
  Transformer.
\newblock \emph{arXiv e-prints} .

\bibitem[{Ruppenhofer et~al.(2016)Ruppenhofer, Ellsworth, Schwarzer-Petruck,
  Johnson, Baker, and Scheffczyk}]{ruppenhofer2006framenet}
Ruppenhofer, J.; Ellsworth, M.; Schwarzer-Petruck, M.; Johnson, C.~R.; Baker,
  C.; and Scheffczyk, J. 2016.
\newblock FrameNet II: Extended theory and practice .

\bibitem[{Shen and Lapata(2007)}]{shen2007using}
Shen, D.; and Lapata, M. 2007.
\newblock Using semantic roles to improve question answering.
\newblock In \emph{Proceedings of the 2007 joint conference on empirical
  methods in natural language processing and computational natural language
  learning (EMNLP-CoNLL)}, 12--21.

\bibitem[{Singh et~al.(2002)Singh, Lin, Mueller, Lim, Perkins, and
  Zhu}]{singh2002open}
Singh, P.; Lin, T.; Mueller, E.~T.; Lim, G.; Perkins, T.; and Zhu, W.~L. 2002.
\newblock Open mind common sense: Knowledge acquisition from the general
  public.
\newblock In \emph{OTM Confederated International Conferences" On the Move to
  Meaningful Internet Systems"}, 1223--1237. Springer.

\bibitem[{Speer, Chin, and Havasi(2016)}]{speer2016conceptnet}
Speer, R.; Chin, J.; and Havasi, C. 2016.
\newblock Conceptnet 5.5: An open multilingual graph of general knowledge.
\newblock \emph{arXiv preprint arXiv:1612.03975} .

\bibitem[{Stevenson and Lindberg(2010)}]{noad}
Stevenson, A.; and Lindberg, C.~A., eds. 2010.
\newblock \emph{New Oxford American Dictionary, Third Edition}.
\newblock Oxford University Press.

\bibitem[{Swayamdipta et~al.(2017)Swayamdipta, Thomson, Dyer, and
  Smith}]{swayamdipta2017frame}
Swayamdipta, S.; Thomson, S.; Dyer, C.; and Smith, N.~A. 2017.
\newblock Frame-semantic parsing with softmax-margin segmental rnns and a
  syntactic scaffold.
\newblock \emph{arXiv preprint arXiv:1706.09528} .

\end{thebibliography}

\end{document}